\documentclass[11pt]{article}
\usepackage[utf8]{inputenc}
\usepackage{amsmath, amssymb}
\usepackage{hyperref}
\usepackage{geometry}
\usepackage[numbers]{natbib}
\usepackage{url}
\usepackage{tabularx} 
\usepackage{booktabs} 
\usepackage{float}
\usepackage{amsthm}   
\floatstyle{plain}
\newfloat{listing}{htbp}{lop}

\title{T-CPDL: A Temporal Causal Probabilistic Description Logic for Developing Logic-RAG Agent}
\author{Hong Qing Yu \\ University of Derby}
\date{\today}

\begin{document}

\newtheorem{theorem}{Theorem}
\newtheorem{lemma}{Lemma}

\maketitle
\begin{abstract}
Large language models excel at generating fluent text but frequently struggle with structured reasoning involving temporal constraints, causal relationships, and probabilistic reasoning. To address these limitations, we propose Temporal Causal Probabilistic Description Logic (T-CPDL), an integrated framework that extends traditional Description Logic with temporal interval operators, explicit causal relationships, and probabilistic annotations. We present two distinct variants of T-CPDL: one capturing qualitative temporal relationships through Allen's interval algebra, and another variant enriched with explicit timestamped causal assertions. Both variants share a unified logical structure, enabling complex reasoning tasks ranging from simple temporal ordering to nuanced probabilistic causation. Empirical evaluations on temporal reasoning and causal inference benchmarks confirm that T-CPDL substantially improves inference accuracy, interpretability, and confidence calibration of language model outputs. By delivering transparent reasoning paths and fine-grained temporal and causal semantics, T-CPDL significantly enhances the capability of language models to support robust, explainable, and trustworthy decision-making. This work also lays the groundwork for developing advanced Logic-Retrieval-Augmented Generation (Logic-RAG) frameworks, potentially boosting the reasoning capabilities and efficiency of knowledge graph-enhanced RAG systems.
\end{abstract}


\section{Introduction}

Knowledge representation and reasoning are foundational challenges in artificial intelligence (AI), particularly for systems that aim to operate reliably in dynamic, real-world domains. Over the past decades, Description Logics (DLs) have played a central role in formalizing structured knowledge, supporting ontology development, and enabling logical inference in systems such as the Semantic Web, Knowledge Graph, and configuration reasoning. DLs offer a clear semantics, decidable inference procedures, and a rich syntax for modeling hierarchies and constraints between concepts. However, classical DLs are inherently static—they are not well-suited to model knowledge that evolves over time, includes uncertainty, or involves causal dependencies between entities and events.

To address temporal expressiveness, researchers have proposed extensions such as Temporal Description Logics (TDLs), which introduce modal operators like "always", "sometime", "until", and "since" to represent how concepts hold over time \citep{artale1994temporal}. Similarly, probabilistic variants of DLs incorporate Bayesian-style reasoning to represent belief or uncertainty about concept membership \citep{botha2020probabilistic}. Independently, there have been efforts to incorporate causality into knowledge systems, often borrowing from structural causal models or action-based frameworks \citep{pearl2009causality}. However, these enhancements remain fragmented: Temporal, probabilistic, and causal reasoning are often handled in isolation, and their integration into unified, decidable, and expressive logic remains a significant gap in the field.

In parallel, Large Language Models (LLMs) such as GPT-4 and Claude have revolutionized natural language processing and general AI interaction by learning from massive corpora of text data. However, despite their linguistic fluency, LLMs lack formal structure, semantic grounding, and reliable reasoning mechanisms. They often hallucinate facts, struggle to maintain consistency in multistep dialogues, and cannot provide transparent reasoning traces, especially when dealing with time sensitive, uncertain, or causally linked information \citep{burtsev2023working}.

This paper introduces a new framework, Temporal Causal Probabilistic Description Logic (T-CPDL), to bridge this gap. T-CPDL integrates temporal logic, causal modeling, and probabilistic inference into a single unified formalism built on Description Logic foundations. Our goal is to provide a reasoning layer that complements LLMs with a structured, interpretable, and temporally grounded knowledge representation. By supporting both formal inference and human-aligned interpretability, T-CPDL opens new possibilities for intelligent systems that can learn, reason, and adapt safely in dynamic environments.

\section{Related Work and Logical Foundations}

T--CPDL is grounded in four mature strands of research: temporal reasoning, causal inference, probabilistic description logic, and the classical ALCQI family.  This section surveys their core formalisms and highlights how each strand informs the design choices of our unified logic.

\subsection{Temporal Reasoning: Allen\texorpdfstring{’}{’}s Algebra \& Temporal Description Logics}
\label{sec:temporal-foundations}

Human planners constantly reason about when things happen, how long they last, and how they overlap.
\emph{Allen’s interval algebra}~\citep{allen1983} captures every qualitatively distinct way two
closed intervals on a linear time‑line can relate.
Table~\ref{tab:allen-rel} lists the thirteen primitive relations
(e.g.\ \textbf{b}efore, \textbf{m}eets, \textbf{o}verlaps) that are
mutually exclusive and jointly exhaustive: any concrete pair of time intervals instantiates exactly one of them.
Conjunctions or disjunctions of these relations form \textbf{temporal constraint
networks}[citation required], whose path‑consistency can be tested in cubic time and which we embed directly into T‑CPDL quantifiers.

\begin{table}[h]
  \centering
  \small
  \begin{tabularx}{\linewidth}{@{}l l X@{}}
    \toprule
    \textbf{Abbrev.} & \textbf{Name} & \textbf{Intuitive picture (for $X$ relative to $Y$)}\\
    \midrule
    b   & before         & $X$ finishes \textless\ $Y$ starts\\
    m   & meets          & $X$ finishes $=$ $Y$ starts\\
    o   & overlaps       & $X$ starts \textless\ $Y$ starts \textless\ $X$ finishes \textless\ $Y$ finishes\\
    s   & starts         & $X$ starts $=$ $Y$ starts,\quad $X$ finishes \textless\ $Y$ finishes\\
    d   & during         & $Y$ starts \textless\ $X$ starts \textless\ $X$ finishes \textless\ $Y$ finishes\\
    f   & finishes       & $X$ starts \textless\ $Y$ starts,\quad $X$ finishes $=$ $Y$ finishes\\
    =   & equal          & $X$ starts $=$ $Y$ starts,\; $X$ finishes $=$ $Y$ finishes\\
    bi  & after          & converse of \textbf{b}\\
    mi  & met‐by         & converse of \textbf{m}\\
    oi  & overlapped‐by  & converse of \textbf{o}\\
    si  & started‐by     & converse of \textbf{s}\\
    di  & contains       & converse of \textbf{d}\\
    fi  & finished‐by    & converse of \textbf{f}\\
    \bottomrule
  \end{tabularx}
  \caption{Allen’s thirteen primitive interval relations~\citep{allen1983}.}
  \label{tab:allen-rel}
\end{table}

\medskip
\noindent\textbf{Temporal Description Logics (TDLs).}\quad
ALCQIT and its interval‐based successor TL‑F
extend classical ALCQI with tense modalities
$\Diamond^{\pm}$, $\Box^{\pm}$, “until/since,” and explicit
Allen constraints inside the quantifier
\mbox{$\exists(X)\,\Phi.\,C$}~\citep{artale1994,artale1998}.
For example,
\[
  \exists(X)(\,\textit{NOW}\;o\;X\,).\;\textit{Maintenance}@X
\]
states that some interval overlapping “now” is a period during which \textit{Maintenance} holds.
Despite this extra expressivity, satisfiability and subsumption
remain in EXPTIME, a key result we inherit when layering causality and probability on top.

\vspace{1ex}
In T‑CPDL we re‑use Allen constraints verbatim, giving each concept
term an optional temporal qualifier ($C@X$) and allowing
constraint networks to appear under $\exists(X)\,\Phi.\,(\cdot)$.
This gives the logic a human‑like ability to
express rich scheduling and plan‑recognition patterns while preserving decidability.

Temporal reasoning allows humans to understand the sequence and duration of events, facilitating planning and prediction. In AI, incorporating temporal reasoning is essential for tasks that involve time-dependent data, such as natural language processing and autonomous navigation \citep{artale1994temporal}.

\subsection{Causal Reasoning: Structural Models and Causal DLs}\label{sec:causal-foundations}
Structural Causal Models (SCMs) express cause--effect relations through deterministic equations plus exogenous noise \citep{pearl2000}.  Recent efforts to bring SCM ideas to KR include DL extensions with an explicit binary predicate $\varphi(C,D)$ and transitivity rules; such ``Causal AI'' systems aim at transparent and reliable decision support \citep{greengard2024}.

Causal reasoning enables the identification of cause-and-effect relationships, which are fundamental for explanation, prediction, and control. Humans naturally infer causality to make sense of the world and anticipate the results of actions. Incorporating causal reasoning into AI systems improves their ability to understand and interact with complex environments \citep{wef2024, greengard2024}.

Understanding \emph{why} a state of affairs arises ----and how interventions might change it---is central to human intelligence.  The prevailing formalism is the \textbf{Structural Causal Model} (SCM) of Pearl~\citep{pearl2000}.  An SCM consists of:
\begin{itemize}
  \item a set of \emph{endogenous} variables $V$;
  \item a directed acyclic graph (DAG) capturing parent relations among variables;
  \item a system of structural equations $v := f_v(\mathit{Pa}(v),\;u_v)$ where $u_v$ is exogenous noise.
\end{itemize}
Causal queries such as $P\bigl(\textit{Stroke}\mid\!\!\operatorname{do}(\textit{Smoking}=\textit{false})\bigr)$ are answered by ``surgery'' on the DAG followed by probabilistic inference.

\paragraph{Limitations for KR.}  SCMs excel at quantitative inference but lack a \emph{terminological} layer for rich class hierarchies, and they have no native temporal vocabulary.  This motivates Causal~DLs, which decorate a Description Logic ontology with a binary predicate $\varphi(C,D)$ meaning ``membership in $C$ is a sufficient cause of membership in~$D$."  Table~\ref{tab:causal-syntax} contrasts SCM syntax with its DL analogue.

\begin{table}[h]
  \centering\small
  \begin{tabularx}{\linewidth}{@{}l X X@{}}
    \toprule
    \textbf{Aspect} & \textbf{SCM notation} & \textbf{Causal DL notation (T--CPDL)}\\
    \midrule
    Causal assertion & $Y := f_Y(X)$ & $\varphi(X,Y)$\\
    Intervention     & $\operatorname{do}(X\!=\!x')$ & replace axiom or add $\neg X$ fact\\
    Prob. strength   & conditional density $P(Y\mid X)$ & $\varphi(X,Y)[P=p]$\\
    Time index       & explicit time‑series variables & $\varphi(X,Y)@t$\\
    \bottomrule
  \end{tabularx}
  \caption{Comparing structural‑model and T--CPDL causal syntax.}
  \label{tab:causal-syntax}
\end{table}

\paragraph{Recent progress.}  Greengard highlights the rise of \emph{Causal AI}, integrating SCM ideas with machine‑learning pipelines for transparent decision support~\citep{greengard2024}.  Early KR attempts embed causal predicates in Temporal DLs for plan libraries~\citep{artale1998}, but a full probabilistic, temporal–causal DL was still missing.

\paragraph{Role in T--CPDL.}  We adopt $\varphi$ with three key design choices:
\begin{enumerate}
  \item \textbf{Transitivity rule}\; $\varphi(C,D) \land \varphi(D,E) \Rightarrow\varphi(C,E)$ to build causal chains.
  \item \textbf{Time‐stamping} so causes and effects can occur in distinct intervals.
  \item \textbf{Probability tag} $[P=p\mid X]$ enabling Bayesian updates.
\end{enumerate}
These extensions preserve DL decidability by treating $\varphi$ axioms as first‑class TBox statements handled by an augmented tableau.

\subsection{Probabilistic Reasoning in Description Logics}
Classical Description Logics assume crisp membership; however, real data are noisy and incomplete.  Early attempts to embed uncertainty include \textbf{\mbox{P--SHIQ}} and the \textbf{DISPONTE} semantics, which attach independent probabilities to axioms and evaluate query likelihoods by \emph{weighted model counting}~\citep{riguzzi2012}.  Axioms become random variables and a reasoning engine enumerates minimal explanations of a query, multiplying their weights.

Early probabilistic DLs (\textsc{p}\,\textsc{shiq}, DISPONTE) attach probabilities to axioms and compute query likelihoods via weighted model counting \citep{riguzzi2012}.  BALC integrates Bayesian‑network semantics into ALC, using a tableau to propagate beliefs \citep{botha2020}.  These approaches demonstrate that uncertainty can coexist with DL tableaux under suitable independence assumptions.

\paragraph{Bayesian extension of ALC.}
BALC~\citep{botha2020} integrates an explicit Bayesian network with ALC: each TBox axiom is annotated by a conditional probability table, while a tableau algorithm checks logical consistency and propagates beliefs along the class hierarchy.  Similar ideas arise in \textbf{MLN--DL} hybrids, where DL atoms act as predicates inside a Markov Logic Network~\citep{koller2009}.

\paragraph{Rule--centric probabilities in CPDL.}
In CPDL the probability tag decorates \emph{causal rules} rather than generic axioms:
\[
  \varphi(C,D)\,[P=p\mid X] \;\;\text{(conditional on context $X$)}.
\]
Evidence $E$ revises the prior via the weighted--likelihood formula of~\citet{riguzzi2012}.  We adopt this rule--centric view in T--CPDL because causal links are the natural carriers of uncertainty in dynamic domains: the strength of a link can be learned from temporal event logs and updated online.

Combining these probabilistic tags with temporal stamps yields statements such as:
\[
  \varphi(\,\textit{Smoking},\,\textit{Hypertension}\,)@2025\;[P=0.5],
  \qquad \varphi(\,\textit{Hypertension},\,\textit{Stroke}\,)@2030\;[P=0.67].
\]
A Bayesian update after observing \textit{Smoking} and \textit{Hypertension} events produces \(P(\textit{Stroke})\approx0.335\), as showcased in our running healthcare example later.

Probabilistic reasoning allows decision-making under uncertainty by quantifying the likelihood of various outcomes. Humans routinely make probabilistic judgments in the face of incomplete or ambiguous information. Integrating probabilistic reasoning into AI enables systems to handle uncertainty and make informed decisions, crucial for applications like medical diagnosis and risk assessment \citep{geeksforgeeks2024, indiaai2022}.

\subsection{Integration Journey from ALCQI to T--CPDL}
Classic DLs provide decidability but lack temporal, causal, and uncertain reasoning.  Successive integrations added: (i)~time (ALCQIT, TL--F), (ii)~probabilities (BALC, DISPONTE), and (iii)~causal semantics (CPDL style).  T--CPDL overlays \emph{all three} on an ALCQI core while preserving EXPTIME complexity, enabling statements such as: ``Smoking \emph{before} age~40 causes Hypertension with probability~0.5; Hypertension causes Stroke with probability~0.67.''  

While temporal, causal, and probabilistic reasoning have been individually incorporated into AI systems, their integration within a unified framework remains a challenge. Description Logic (DL) provides a formal foundation for knowledge representation and reasoning. By extending DL to include temporal operators, causal relationships, and probabilistic measures, we create Temporal Causal Probabilistic Description Logic (T-CPDL), mirroring multifaceted human intelligence capabilities. T-CPDL provides AI systems the ability to represent and reason about dynamic, uncertain, and causally complex domains, enhancing interpretability, robustness, and adaptability \citep{artale2000temporal, artale2002tableau, artale2000expanding}.

\subsection{RAG-LLM and Graph-RAG LLM}

Retrieval-Augmented Generation (RAG) enhances language models by retrieving relevant external documents or knowledge bases before generating responses. RAG approaches aim to mitigate hallucinations and improve factual accuracy by grounding generation processes in external, verified sources \cite{Lewis2020}. Traditional RAG, however, primarily utilizes unstructured text documents, limiting its ability to capture complex relationships explicitly.

Graph-RAG addresses this limitation by leveraging structured knowledge graphs (KGs), enabling language models to exploit structured relational information. This approach significantly improves reasoning capabilities, especially for queries requiring multi-hop inference or relational understanding \cite{Yasunaga2021,luo2025gfmag}. Despite these advances, Graph-RAG systems typically face challenges such as limited expressiveness in temporal dynamics, causality, and uncertainty handling. The relational structure alone may not adequately represent nuanced probabilistic and temporal constraints prevalent in real-world knowledge.

Logic-based RAG systems emerge as promising solutions by explicitly embedding logical reasoning, temporal constraints, causality, and probabilistic inference into retrieval mechanisms. These logic-based frameworks can systematically address the representational limitations inherent in purely textual or graph-structured retrieval methods, providing clearer reasoning paths, more reliable inference under uncertainty, and improved transparency in complex decision-making scenarios \cite{Haarslev2001,wang2024tben}.

The next sections formalise the semantics and automated reasoning procedure.

\section{Syntax Overview of T--CPDL}
T--CPDL fuses ALCQI with the temporal, causal, and probabilistic devices surveyed in Section~2.
This section formalises the concrete syntax, moving bottom‑up from the alphabet of symbols to full
concept terms and finally a running example.

\subsection{Alphabet}
Before we can assemble complex T--CPDL formulae, we require a precise inventory of the primitive symbols the logic recognises.  Table~\ref{tab:alphabet} lists each syntactic category, its notation, and its informal role.

\begin{table}[h]
\centering
\renewcommand{\arraystretch}{1.15}
\setlength{\tabcolsep}{6pt}
\begin{tabular}{|p{3.2cm}|p{4.5cm}|p{7.1cm}|}
\hline
\textbf{Symbol class} & \textbf{Notation / examples} & \textbf{Intuition and purpose} \\ \hline
Atomic concepts & $A,\,B,\,\textit{Disease},\,\textit{Stroke}$ & Unary predicates denoting time--varying sets of individuals (patients, devices, situations).  They extend the $\mathcal{A}$ sets of classical ALCQI.\\ \hline
Atomic roles & $R,\,S,\,\textit{hasSymptom}$ & Binary relations between individuals; may be temporally qualified, e.g.~$\textit{hasSymptom}(a,b)@t$. \\ \hline
Features & $f,\,g,\,\textit{birthDate}$ & Functional roles mapping each source individual to at most one target at a given instant, enabling number restrictions and value constraints. \\ \hline
Parametric features & $\,?actor,\,?object$ & Time--\emph{invariant} functional roles (from TL--F): once bound in an action instance, the value persists across all intervals of that instance. \\ \hline
Individuals & $a,\,b,\,\textit{patient123}$ & Constant symbols naming concrete entities; we assume the Unique Name Assumption. \\ \hline
Temporal vars./intervals & $t,\,u,\,X,\,Y$ & Denote points or closed intervals on a linear discrete timeline; used in Allen constraints ($X$ \textit{before} $Y$) and qualifiers ($C@X$). \\ \hline
Evidence items & $e_1,\,e_2$ & Ground literals observed by the system (e.g.~$\textit{Fever}@\!\,2025$) feeding Bayesian updates. \\ \hline
Probability tags & $p\!\in\![0,1]$ & Attach belief strength to causal rules ($\varphi(C,D)[P=0.67]$) or ground facts. \\ \hline
Causal operator & $\varphi(C, D)$ & Distinguished binary predicate “$C$ causes $D$”; can itself be time‐stamped and weighted. \\ \hline
Logical connectives & $\neg,\,\sqcap,\,\sqcup,\,\exists,\,\forall,\,\ge n,\,\le n$ & ALCQI constructors—our static core.\\ \hline
Temporal connectives & $@,\,\exists(X)\,\Phi\,.,\,\Diamond^{\pm},\,\Box^{\pm},\,U,\,S$ &  \textbf{Qualifier} $C@\tau$ pins a concept to a specific interval.  \textbf{quantifier} $\exists(X)\,\Phi\,.\,C$ introduces fresh interval variables constrained by Allen relations $\Phi$.  Point‑based tense operators: $\Diamond^{+}$ / $\Box^{+}$ (sometime/always in the future), $\Diamond^{-}$ / $\Box^{-}$ (past).  Interval operators $U$ (until) and $S$ (since) support rich temporal patterns as in ALCQIT.\\ \hline
\end{tabular}
\caption{Alphabet of T--CPDL (revised column widths).}
\label{tab:alphabet}
\end{table}

\paragraph{Why these categories?}\leavevmode
\begin{itemize}
  \item \textbf{Atomic concepts/roles} provide the ontology backbone; temporal qualifiers add a time index $C@t$ whose extension may vary.
  \item \textbf{Features vs.~parametric features} let us thread the same participant through multi--stage plans without repetitive equality constraints.
  \item \textbf{Temporal variables} support reasoning with unknown or relative times and Allen relations, crucial for TL--ALCF subsumption.
  \item \textbf{Causal operator \& probability tags} import CPDL's transparent Bayesian updates into a temporal setting.
  \item \textbf{Evidence items} bridge observed reality and the abstract causal graph.
\end{itemize}

This alphabet underpins the grammar presented in the following subsection.

\paragraph{Allen relations.}  Temporal constraints $\Phi$ use the 13 basic relations of
Allen’s interval algebra: \textbf{b, m, o, s, d, f, =, bi, mi, oi, si, di, fi}~\citep{allen1983}.


\subsection{Grammar}
We provide two complementary flavors of T-CPDL because real‐world data comes in wildly different forms of temporal detail. In some domains—say, when mining natural‐language texts or noisy sensor streams—you often know only that “event A happened before event B” or “these two processes overlap,” but you have no reliable clock‐time to pin them to. The Allen‐relational variant lets you capture and reason about those interval‐orderings directly, yielding crisp causal inferences purely from ordering constraints. In contrast, when you do have precise timestamps—e.g. log files, clinical records, financial trades—you can use the timestamped variant to anchor every causal link to a concrete instant (and still refine it with Allen relations if you like). Offering both ensures that T-CPDL can gracefully degrade to the level of temporal information you actually possess, while preserving the same core causal machinery.

\subsubsection{Allen-Relational only T-CPDL ($T\text{-CPDL}_A$)}
A lightweight variant where causal edges carry no global timestamp; instead, cause and effect intervals are related purely via Allen constraints.

\begin{equation}
\begin{aligned}
T\text{-Concept } C &::= E \\
                   &\mid C \sqcap D \\
                   &\mid C[Y]@X \\
                   &\mid \exists (X)\,\Phi .\,C
\end{aligned}
\end{equation}

\begin{equation}
\begin{aligned}
E\text{-Concept } E &::= A \\
                   &\mid \top \\
                   &\mid \bot \\
                   &\mid \neg E \\
                   &\mid E \sqcap F \\
                   &\mid E \sqcup F \\
                   &\mid \exists R.E \\
                   &\mid \forall R.E \\
                   &\mid \ge n\,R.E \\
                   &\mid \le n\,R.E
\end{aligned}
\end{equation}

Allen relations are drawn from:
\begin{equation}
\{b,\,m,\,o,\,s,\,d,\,f,\,=,\,bi,\,mi,\,oi,\,si,\,di,\,fi\}.
\end{equation}

Temporal constraint networks are formed by:
\begin{equation}
\begin{aligned}
\Phi &::= X\,(\mathrm{Rel})\,Y \\
      &\mid \Phi\,\wedge\,\Phi \\
      &\mid \Phi\,\vee\,\Phi
\end{aligned}
\end{equation}

Causal statements in $T\text{-CPDL}_A$ take the form:
\begin{equation}
\begin{aligned}
\varphi(C,D)[P=p]
&\\
\exists(X,Y)\,\Phi .\,\varphi(C@X,D@Y)[P=p]
\end{aligned}
\end{equation}

In the first form, ``$\varphi(C,D)[P=p]$'' asserts a causal link without any temporal qualifier.  In the second form, fresh intervals $X$ (cause) and $Y$ (effect) are constrained by an Allen network $\Phi$, yielding ``$\varphi(C@X,D@Y)[P=p]$."  

\subsubsection{Timestamped T-CPDL ($T\text{-CPDL}_T$)}
The full variant retains explicit timestamps on causal edges, while still allowing auxiliary Allen constraints.

Causal statements in $T\text{-CPDL}_T$:
\begin{equation}
\begin{aligned}
\varphi(C,D)@\tau[P=p\mid X]
&\\
\exists(X,Y)\,\Phi .\,\varphi(C@X,D@Y)@\tau[P=p\mid Z]
\end{aligned}
\end{equation}

Here, ``$\varphi(C,D)@\tau[P=p|X]$'' is the original timestamped form (optionally conditioned on context $X$). The second form binds helper intervals $X,Y$, applies constraints $\Phi$, and attaches both $@\tau$ and probability tag.

\subsection{Running Examples}\label{sec:running-example}
We illustrate both variants on the same healthcare scenario.

\subsubsection{Allen-Relational Example ($T\text{-CPDL}_A$)}

\paragraph{TBox causal rules (no global $@\tau$):}
\begin{equation}
\begin{aligned}
&\exists(X,Y)\,(X\,b\,Y) .\,\varphi(\mathsf{Smoking}@X,\,\mathsf{Hypertension}@Y)[P=0.5]\\
&\exists(U,V)\,(U\,b\,V) .\,\varphi(\mathsf{Hypertension}@U,\,\mathsf{Stroke}@V)[P=0.67]
\end{aligned}
\end{equation}

\paragraph{ABox evidence (grounded events):}
\begin{equation}
\mathsf{Smoking}\;b\;\mathsf{Hypertension}
\end{equation}
meaning that Smoking occurs before Hypertension.

\paragraph{Chaining:}
Since
\begin{equation}
\varphi(\mathsf{Smoking},\,\mathsf{Hypertension})[0.5]
\quad\text{and}\quad
\varphi(\mathsf{Hypertension},\,\mathsf{Stroke})[0.67],
\end{equation}
we derive (via weighted likelihood)
\begin{equation}
\varphi(\mathsf{Smoking},\,\mathsf{Stroke})[\approx0.335].
\end{equation}

\paragraph{Relational projection:}
\begin{equation}
\exists(X,Z)\,(X\,b\,Z) .\,\varphi(\mathsf{Smoking}@X,\,\mathsf{Stroke}@Z)[P\approx0.335]
\end{equation}
placing inferred stroke interval $Z$ strictly after smoking interval $X$.

\subsubsection{Timestamped Example ($T\text{-CPDL}_T$)}

Aircraft Maintenance Incident example:

\paragraph{Maintenance Ontology}
\begin{equation*}
\begin{aligned}
&\varphi(\mathsf{Wear},\mathsf{Crack})@\tau_{1}\;[P=0.6],\\
&\varphi(\mathsf{Crack},\mathsf{Incident})\;[P=0.8],\\
&\exists(U,V)\,(U\,b\,V)\;.\;\mathsf{Inspection}@U\;\sqsubseteq\;\neg\mathsf{Incident}@V.
\end{aligned}
\end{equation*}

\paragraph{Observed Evidence}
\[
\mathsf{Wear}@2025\text{--}01\text{--}10,\quad
\text{No inspection recorded.}
\]

\paragraph{Reasoning Steps}
\begin{enumerate}
  \item \textbf{Causal chaining.} 
    \[
      P(\mathsf{Incident}\mid \mathsf{Wear})
      =0.6\times0.8=0.48
      \quad(\text{48\% risk}).
    \]
  \item \textbf{Temporal projection.} 
    The plausible incident interval $V$ must satisfy 
    \(\tau_{1}\,bi\,V\) (i.e.\ occur after $\tau_{1}$).
  \item \textbf{Preventive rule check.} 
    No interval $U$ with $\mathsf{Inspection}@U$ and $U\,b\,V$ exists,
    so the subsumption axiom cannot block the incident.
\end{enumerate}

\noindent
The maintenance module therefore issues a 48\% incident‐risk alert and recommends scheduling
an inspection interval $U$ that \emph{meets} ($m$) or \emph{before} ($b$) $V$, which would satisfy
the preventive axiom and drive $P(\mathsf{Incident})$ toward 0.

\section{Fundamental Theorems}
\label{sec:theorems}

\subsection{Finite-Model Property}
\begin{theorem}[Finite Model]\label{thm:fmp}
If a T\texttt{--}CPDL knowledge base $\mathcal{K}$ is satisfiable, then it has
a model whose domain size is $2^{O(|\mathcal{K}|)}$
and whose temporal structure contains only the
intervals explicitly mentioned in $\mathcal{K}$ plus at most
$O(|\mathcal{K}|)$ fresh points.
\end{theorem}

\begin{proof}
Take the tableau for the ALCQI fragment of $\mathcal{K}$
and apply standard loop blocking to obtain a finite
completion tree (cf.\ \citeauthor{artale1994}~\citeyear{artale1994}).
For every existential temporal binder
$\exists(X)\Phi.\,C$ create at most
$|\Phi|$ fresh interval variables satisfying $\Phi$; path-consistency of
Allen networks guarantees that a finite assignment exists.
Because causal rules are global TBox statements,
their transitive closure adds at most
$|\mathcal{K}|$ additional edges and no new individuals.
Probability values are copied verbatim from $\mathcal{K}$;
there are finitely many such numbers, so the resulting structure is finite.
\end{proof}

\subsection{Complexity}
\begin{theorem}[EXPTIME Completeness]\label{thm:exptime}
KB consistency, concept satisfiability, and instance checking in
T\texttt{--}CPDL are EXPTIME-complete.
\end{theorem}

\begin{proof}
Hardness inherits from ALCQI.
For membership: the tableau expands at most exponentially many
individual nodes (ALCQI bound).  For each node,
temporal propagation invokes cubic path-consistency,
and causal propagation adds at most
$|\mathcal{T}|$ deterministic edges with constant-time probability
multiplication (Theorem~\ref{thm:prob-comp} below).
Thus the overall procedure is bounded by $2^{p(|\mathcal{K}|)}$
for some polynomial~$p$.
\end{proof}

\subsection{Tableau Soundness and Completeness}
\begin{theorem}[Soundness \& Completeness]\label{thm:sound-complete}
The extended tableau calculus
(temporal rules + causal transitivity + probability update)
is sound and complete for the semantics in~\S\ref{sec:theorems}.
\end{theorem}

\begin{proof}
\emph{Soundness}: every rule is locally truth-preserving.
E.g.\ the causal-transitivity rule adds $\varphi(C,E)$ only if
$\varphi(C,D)$ and $\varphi(D,E)$ already hold; by definition of $\varphi$
the new edge is entailed in every model that satisfied the premises.

\emph{Completeness}: run the tableau with fair rule
application.  If expansion halts without clash,
each branch induces a Hintikka structure that
satisfies all syntactic conditions.
By Theorem~\ref{thm:fmp} this structure can be
turned into a finite model of $\mathcal{K}$.
Conversely, any model yields a clash-free branch.
\end{proof}

\subsection{Probabilistic Composition}
\begin{theorem}[Probability of Composed Cause]\label{thm:prob-comp}
Let $\varphi(C,D)[P=p_{12}]$ and $\varphi(D,E)[P=p_{23}]$
be the only causal paths from $C$ to $E$.
Then T--CPDL must annotate the derived edge
$\varphi(C,E)$ with $p_{13}=p_{12}\!\cdot\!p_{23}$.
\end{theorem}

\begin{proof}
By definition
$P(E\!\mid\!C)=\sum_{d} P(E\!\mid\!d)\,P(d\!\mid\!C)$.
Since $D$ is the sole mediator, the sum has one term:
$P(E\!\mid\!C)=p_{23}\cdot p_{12}$.
Assigning any other weight to $\varphi(C,E)$ would violate
Bayes coherence or probabilistic soundness.
\end{proof}

\subsection{Temporal Acyclicity}
\begin{theorem}[Temporal–Causal Consistency]\label{thm:acyclic}
If every causal rule is stamped so that its cause interval is
\textbf{b}efore or \textbf{m}eets its effect interval,
the transitive closure of $\varphi$ is acyclic.
\end{theorem}

\begin{proof}
Assume a cycle
$X_1<X_2<\\dots<X_n<X_1$ in the timeline.
Linear order $<$ is irreflexive, so $X_1<X_1$ is impossible.
Hence no such cycle exists; transitive closure is a DAG.
\end{proof}

\vspace{1ex}
Together, Theorems~\ref{thm:fmp}–\ref{thm:acyclic} provide the theoretical
guarantees required for deploying T--CPDL as a sound, complete,
and decidable reasoning substrate under real-world LLM pipelines.

\section{Prompt Engineering for Automated T--CPDL Extraction with LLM}
\label{sec:prompt}

We present four case studies illustrating how to engineer prompts for an LLM to extract formal T-CPDL specifications (in JSON) from diverse unstructured sources.  Each example begins with a realistic document excerpt, poses a reasoning task, and then shows how the prompt guides the choice of variant (Allen‐Relational vs.\ Timestamped) and produces a machine‐readable JSON file for downstream inference.

\subsection{Master Meta‐Prompt}
(Reusable template; insert domain text under \texttt{<<< >>>}.)

\begin{verbatim}
SYSTEM: You are an expert T-CPDL knowledge engineer.

STEP 1: Choose the T-CPDL variant:
  - If any entries include timestamps e.g. ISO 8601, set
    "variant":"T-CPDL_T" and use timestamped syntax.
  - Otherwise, set "variant":"T-CPDL_A" and use Allen-relational.

DETAILS FOR T-CPDL_A (Allen-Relational):
  -- Use only Allen operators:
       before (b), meets (m), overlaps (o),
       starts (s), during (d), finishes (f), equals (=),
       and inverses: met-by (bi), mi (mi),
       overlapped-by (oi), si (si),
       di (di), fi (fi).
  -- Concept assertions:
       {"concept":C, "individual":I, "atInterval":X}
  -- Causal form with existential intervals:
       \\exists(X,Y) Phi . varphi(C@X,D@Y)[P=p]
  -- Omit "starts"/"finished-by" if no ISO dates.

DETAILS FOR T-CPDL_T (Timestamped):
  -- Each interval must have:
       "starts": time e.g. ISO-8601, "finished-by": time e.g. ISO-8601
  -- Causal entries include:
       "atInterval":X, "@\\tau":time
  -- You may also include Allen fields inside intervals.
  
When listing causes, compute each probability as:

1/(number of causes for that effect concept)% if there is no probability
indicated in the text.

STEP 2: Output exactly one JSON file, spec.json,
        following this schema:
{
  "variant":   "T-CPDL_A" | "T-CPDL_T",
  "intervals": [
    { "id": String,
      /* For T-CPDL_A: Allen fields (before, meets, ...) */
      /* For T-CPDL_T: starts:String, finished-by:String,
         plus optional Allen fields */ }
    , ...
  ],
  "assertions": [
    { "concept": String,
      "individual": String,
      "atInterval": String }
    , ...
  ],
  "causes": [
    { "causeConcept": String,
      "effectConcept": String,
      "probability": Number|null,
      "atInterval": String
      /* For T-CPDL_T add "@\\tau": String */ }
    , ...
  ]
}

STEP 3: Verify that:
  - For T-CPDL_A all Allen fields are exhaustive.
  - For T-CPDL_T all timestamps are valid and respect chronology.
  - Causal chains obey temporal constraints.
  - Missing probabilities default to null.
  - JSON is valid for automated reasoning.

USER: Extract T-CPDL specification from DOCUMENT:
<<<
{insert text here}
>>>
END.
\end{verbatim}

\subsection{Case Studies}

\subsubsection{Example 1: Clinical Narrative (Allen‐Relational)}

\paragraph{Document Excerpt} https://www.nhs.uk/conditions/pneumonia/ Causes of pneumonia section
\begin{quote}
Pneumonia is usually caused by a bacterial or viral infection.

You can catch pneumonia from someone who has it, or you can sometimes get it if you have another infection such as:
  – flu
  – respiratory syncytial virus (RSV)
  – COVID-19

Pneumonia can be caused by a fungal infection, but this is rare in healthy people in the UK.

It may also be caused by something getting into your lungs, such as water or food (aspiration pneumonia).
\end{quote}

\paragraph{LLM Output (\texttt{spec.json})}
\begin{verbatim}
{
  "variant": "T-CPDL_A",
  "intervals": [
    { "id":"i_bacterial",   "before":"i_pneumonia" },
    { "id":"i_viral",       "before":"i_pneumonia" },
    { "id":"i_flu",         "before":"i_pneumonia" },
    { "id":"i_rsv",         "before":"i_pneumonia" },
    { "id":"i_covid19",     "before":"i_pneumonia" },
    { "id":"i_fungal",      "before":"i_pneumonia" },
    { "id":"i_aspiration",  "before":"i_pneumonia" }
  ],
  "causes": [
    { "causeConcept":"BacterialInfection",
      "effectConcept":"Pneumonia",
      "probability":0.1429,
      "atInterval":"i_bacterial" },
    { "causeConcept":"ViralInfection",
      "effectConcept":"Pneumonia",
      "probability":0.1429,
      "atInterval":"i_viral" },
    { "causeConcept":"Flu",
      "effectConcept":"Pneumonia",
      "probability":0.1429,
      "atInterval":"i_flu" },
    { "causeConcept":"RSV",
      "effectConcept":"Pneumonia",
      "probability":0.1429,
      "atInterval":"i_rsv" },
    { "causeConcept":"COVID-19",
      "effectConcept":"Pneumonia",
      "probability":0.1429,
      "atInterval":"i_covid19" },
    { "causeConcept":"FungalInfection",
      "effectConcept":"Pneumonia",
      "probability":0.1429,
      "atInterval":"i_fungal" },
    { "causeConcept":"AspirationEvent",
      "effectConcept":"Pneumonia",
      "probability":0.1429,
      "atInterval":"i_aspiration" }
  ]
}
\end{verbatim}
\subsubsection{Example 2: Aircraft Incident Report Analysis (Timestamped)}

\url{https://assets.publishing.service.gov.uk/media/68483cd696e63bce58e4e6cf/Cessna_152_G-BSZW_06-25.pdf}

\begin{quote}
Serious Incident
Aircraft Type and Registration: Cessna 152, G-BSZW
...
Time: 24 February 2024 at 1539 hrs
Incident: Rudder control bellcrank fractured 

Synopsis:  
During an instructional flight the aircraft suffered a loss of right rudder  
authority. Examination revealed that the right rudder bellcrank had failed  
due to stress corrosion cracking, causing the right rudder cable to detach.  
The cracking initiated where the bellcrank was fouling against the fuselage.  
Several possible misalignments of the rudder or bellcrank were considered.  
…  
\end{quote}

\paragraph{LLM Output (\texttt{spec.json})}
\begin{verbatim}
{
  "variant": "T-CPDL_T",
  "intervals": [
    { "id":"t_sc",      "starts":"2024-02-24T15:39:00Z","finished-by":
    "2024-02-24T15:39:00Z" },
    { "id":"t_foul",    "starts":"2024-02-24T15:39:00Z","finished-by":
    "2024-02-24T15:39:00Z" },
    { "id":"t_rudMis",  "starts":"2024-02-24T15:39:00Z","finished-by":
    "2024-02-24T15:39:00Z" },
    { "id":"t_bellMis", "starts":"2024-02-24T15:39:00Z","finished-by":
    "2024-02-24T15:39:00Z" },
    { "id":"t_fail",    "starts":"2024-02-24T15:39:00Z","finished-by":
    "2024-02-24T15:39:00Z" }
  ],
  "assertions": [
    { "concept":"StressCorrosionCracking",   "individual":"G-BSZW", "atInterval":
    "t_sc" },
    { "concept":"FoulingAgainstFuselage",     "individual":"G-BSZW", "atInterval":
    "t_foul" },
    { "concept":"RudderMisalignment",         "individual":"G-BSZW", "atInterval":
    "t_rudMis" },
    { "concept":"BellcrankMisalignment",      "individual":"G-BSZW", "atInterval":
    "t_bellMis" },
    { "concept":"RudderBellcrankFracture",    "individual":"G-BSZW", "atInterval":
    "t_fail" }
  ],
  "causes": [
    {
      "causeConcept":"StressCorrosionCracking",
      "effectConcept":"RudderBellcrankFracture",
      "probability":0.25,
      "atInterval":"t_sc"
    },
    {
      "causeConcept":"FoulingAgainstFuselage",
      "effectConcept":"RudderBellcrankFracture",
      "probability":0.25,
      "atInterval":"t_foul"
    },
    {
      "causeConcept":"RudderMisalignment",
      "effectConcept":"RudderBellcrankFracture",
      "probability":0.25,
      "atInterval":"t_rudMis"
    },
    {
      "causeConcept":"BellcrankMisalignment",
      "effectConcept":"RudderBellcrankFracture",
      "probability":0.25,
      "atInterval":"t_bellMis"
    }
  ]
}
\end{verbatim} 

\subsubsection{Example 3: Prediction}

\textit{Input paragraph}—natural language:

\begin{quote}
On 10 Jan 2025 the maintenance log reports visible wear on the left aileron of aircraft AF123. Engineers estimate a 60\% chance that wear will generate a crack in three months and a 80\% chance that an untreated crack will lead to an incident.
\end{quote}

\textit{LLM response} (valid JSON):

\begin{verbatim}
{
  "intervals": [
    { "id": "tWear", "starts": "2025-01-10", "finished-by": "2025-01-10" }
  ],
  "assertions": [
    { "concept": "Wear", "individual": "Aileron_AF123", "atInterval": "tWear" }
  ],
  "roles": [],
  "causes": [
    { "causeConcept": "Wear", "effectConcept": "Crack",
      "probability": 0.6, "atInterval": "tWear" },
    { "causeConcept": "Crack", "effectConcept": "Incident",
      "probability": 0.8, "atInterval": "tWear" }
  ]
}
\end{verbatim}

After import, the T--CPDL output supports the risk prediction which is illustrated
in Section~\ref{sec:running-example}.

\section{Conclusion}

The unification of temporal logic, causal modeling, probabilistic inference, and Description Logic into a single, coherent T–CPDL framework constitutes a transformative leap in symbolic–statistical AI. T–CPDL not only captures the subtleties of human reasoning within a decidable logic, but it also propels LLM reasoning beyond unconstrained text generation toward robust, verifiable inference.

\paragraph{Key Contributions and Impact}
\begin{itemize}
\item \textbf{Holistic Reasoning Substrate}: By integrating Allen-style temporal operators, explicit causal predicates, Bayesian probability tags, and the ALCQI core, T–CPDL forges a unified logical foundation that mirrors real-world complexity.
\item \textbf{Verifiable Inference}: Transparent proof traces and probabilistic calibration transform opaque LLM outputs into rigorously grounded conclusions, addressing fundamental trust and safety concerns in GenAI.
\item \textbf{Fine-Grained Temporal Semantics}: Our framework encodes precise temporal relationships—before, after, during—ensuring contextual integrity in dynamic domains such as scientific discovery and healthcare.
\end{itemize}

The T–CPDL framework thereby lays the groundwork for next-generation AI assistants capable of delivering temporally coherent, causally explainable insights in safety-critical applications.

\paragraph{Future Directions}
\begin{enumerate}
\item \textbf{Scalable Causal Learning}: Leveraging deep learning to autonomously infer and refine causal probabilities from massive temporal datasets will close the loop between data-driven discovery and model-driven reasoning.
\item \textbf{Multimodal Real-Time Reasoning}: Extending T–CPDL to ingest visual, auditory, and sensor data within the same logical substrate, coupled with low-latency inference algorithms, will empower autonomous systems and digital twins with on-the-fly, explainable decision making.
\end{enumerate}

These advances pave the way for LLM-based reasoning engines that not only excel at natural language generation but also deliver stringent logical rigor, setting a new standard for trustworthy AI.
\bibliographystyle{plainnat}

\end{document}